\documentclass{article}

\usepackage[preprint]{corl_2026} 

\usepackage{graphicx}
\usepackage{enumitem}
\usepackage{amsfonts}
\usepackage{amsmath}
\usepackage{wrapfig}
\usepackage{booktabs}
\usepackage{caption}
\usepackage{longtable}
\usepackage{array}
\newcolumntype{C}[1]{>{\centering\arraybackslash}p{#1}}
\usepackage{xcolor}

\definecolor{darkgreen}{rgb}{0.447,0.831,0.733}
\definecolor{darkred}{rgb}{0.6,0,0}
\definecolor{darkblue}{rgb}{0.447,0.545,0.831}

\hypersetup{
    bookmarks=true,
    colorlinks=true,
    pdfpagemode=UseNone,
    citecolor=darkgreen,
    linkcolor=darkblue,
    urlcolor=darkblue,
    filecolor=darkgreen
}

\title{TAC-LOCO: Unified Whole-Body Control for Quadrupedal TACtile-Informed LOCO-Manipulation}

\author{
\normalfont
\parbox{\textwidth}{
\centering
{\bfseries
Muqun Hu$^{1,*}$,
Yuhao Zhou$^{1,*}$,
Kabir Ray Malik$^{1}$,
Chi Lin$^{1}$\\
Won Suk Lee$^{2}$,
Yu She$^{1}$,
Yan Gu$^{1,\dagger}$
}\\
\vspace{0.3em}
{\normalfont $^{1}$Purdue University}
{\normalfont $^{2}$University of Florida}\\
{\normalfont $^{*}$Equal contribution. \quad $^{\dagger}$Corresponding author.}
}
}

\begin{document}
\maketitle


\begin{abstract}
    Dynamic loco-manipulation requires legged robots to coordinate whole-body motion while maintaining stable physical interaction with grasped objects under uncertain external forces. While tactile sensing has been widely studied for robotic manipulation, its role in dynamic whole-body control remains largely unexplored. Existing works without tactile feedback commonly grasp firmly rather than regulate the grasp according to the interaction. We propose TAC-LOCO, a tactile-augmented unified reinforcement learning framework that encodes tactile array observations from compliant grippers into a compact latent representation and joins it with proprioception for unified control of the legs, arm, and gripper. With effective grasp stability reward design, the policy learns to simultaneously track body velocity and end-effector trajectories, moderate grasp force, and prevent object slip under both gradual load changes and sudden release events. We deploy the policy zero-shot on a Unitree Go2 with an Interbotix WidowX 250 arm and tactile gripper, demonstrating dynamic tactile-informed loco-manipulation under varying external interactions, achieving a 47\% reduction in grasping force and an object drop rate of less than 1\%.
\end{abstract}


\begin{figure}[htbp]
    \centering
    \includegraphics[width=1.0\linewidth]{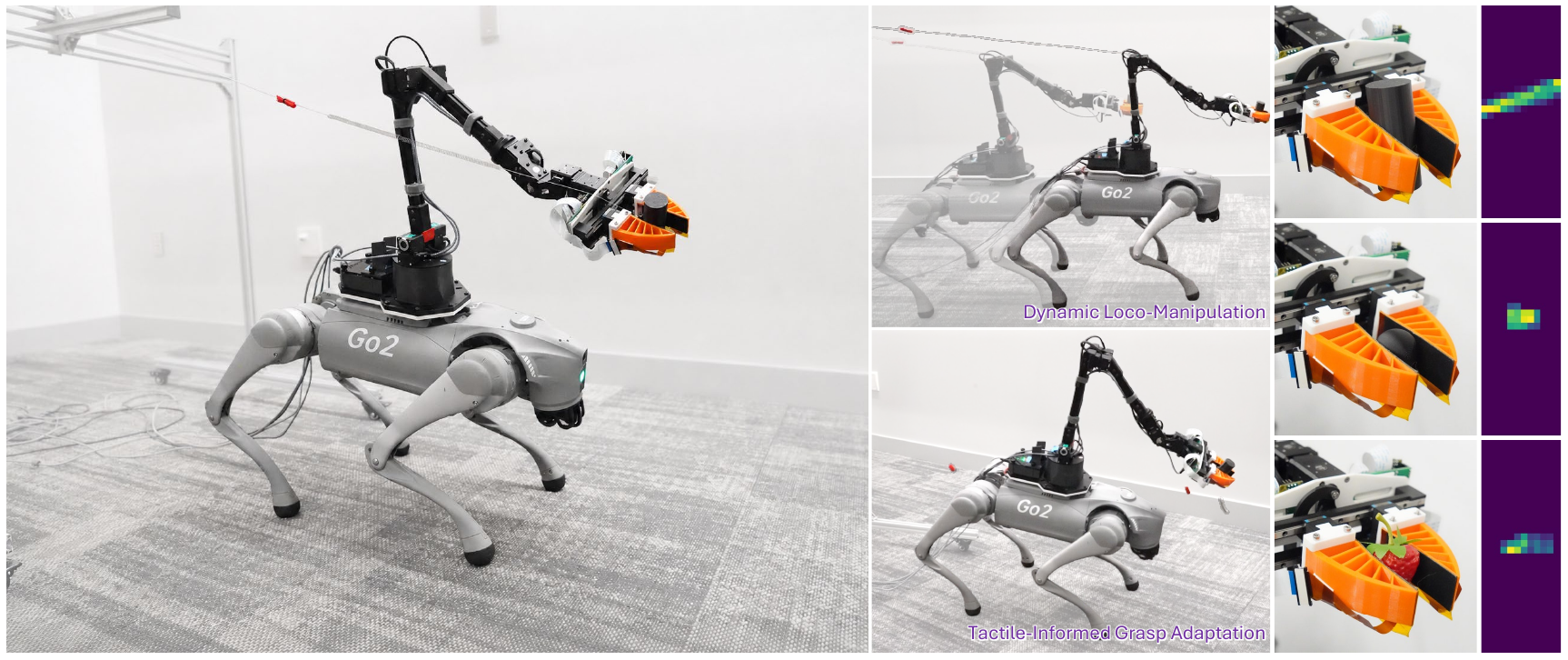}
    \caption{\textbf{We present a unified tactile-informed whole-body control policy for dynamic loco-manipulation}, reducing grasping force and preventing object slip during unknown and time-varying external forces. The policy can accommodate a variety of object shapes and sizes.}
    \label{fig:teaser}
    
\end{figure}

\section{Introduction}
    Legged robots have rapidly evolved from robust locomotion platforms~\citep{hwangbo2019learning, iqbal_provably_2020, zhuang2023robot,iqbal2023mechatronics,gao2025time,he2025invariant,misenti2025experimental,tay2026hybridmimic,mandali2026proprioceptive} into mobile manipulation systems capable of interacting with complex environments~\citep{fu2022deep, iyer2026vectorizing, hu2025towards,gu2025evolution,chang2026survey}. Recent advances in whole-body control (WBC) and reinforcement learning (RL) have enabled quadruped robots with arms to coordinate locomotion and manipulation, extending their reachable workspace and supporting tasks such as reaching, picking, pushing, pressing, and object transport during locomotion~\citep{lin2025locotouch}. This progress opens the door to dynamic contact-rich tasks, where the robot must not only track an end-effector (EE) trajectory but also maintain stable physical interaction with the environment.

    While whole-body loco-manipulation of unconstrained objects in static or quasi-static environments has been extensively studied \citep{liu2024visual, liu2026mlm, 11155187, jung2026dynamic, portela2024learning, qiu2025wildlma, hou2025efficient, 10925884}, many real-world tasks remain challenging when objects are subject to unknown and time-varying external forces. This difficulty is particularly pronounced in contact-rich manipulation, where successful execution requires the robot to continuously regulate in-hand contact forces in response to changing external loads. For instance, when pulling fruit from a stem, the resisting force may gradually increase as the stem deforms and then disappear abruptly upon detachment. Without sensing these evolving interaction forces, the robot may fail to adjust its grasp strength or whole-body motion in time, leading to object slip, excessive contact force, or unstable manipulation. 

    Despite the importance of interaction forces, existing legged loco-manipulation policies typically rely on implicit modeling to estimate external forces applied to robot body links, without directly observing in-hand grasp forces~\citep{zhi2025learning, portela2024learning, xu2025facet}. This may limit the robot’s ability to regulate contact in dynamic manipulation tasks, since the lack of direct observation of object contact state may lead to slower reactions to object slip. As a result, these methods typically prescribe the gripper state or grasping mode manually, rather than learning dynamic, task-adaptive gripper actions. On the other hand, tactile sensors provide direct access to rich contact information and have been widely studied in robotic manipulation~\citep{zhou2026tactile, hogan2020tactile,zhou2024hand}; however, they remain underexplored as policy inputs for dynamic whole-body loco-manipulation control. In contrast, TAC-LOCO integrates tactile observations from compliant grippers into \textbf{one unified whole-body policy} that learns gripper actions together with whole-body motion, addressing the coupled nature of dynamic object movement and grasping regulation, enabling the robot to adapt its grasp in response to changing interaction forces during simultaneous loco-manipulation.

    The main contributions of this work are summarized as follows:
    \begin{enumerate}[leftmargin=*]
        \item [(a)] We propose TAC-LOCO, a tactile-informed unified RL framework that jointly controls the legs, arm, and gripper for dynamic whole-body loco-manipulation under unknown time-varying external forces.
        \item [(b)] We design an MDP formulation that integrates compact tactile latent observations with proprioception, enabling the policy to infer grasp state and adapt gripper actions online, reducing the grasping force by $47\%$ with an object drop rate of less than $1\%$.
        \item [(c)] We validate the framework in simulation and zero-shot hardware experiments (Fig. \ref{fig:teaser}), demonstrating a $90\%$ success rate during dynamic loco-manipulation and generalizability to various object shapes. To the best of our knowledge, this is the first whole-body control framework for dynamic legged loco-manipulation that jointly unifies tactile sensing, gripper regulation, and whole-body policy learning.
    \end{enumerate}

\section{Related Works}
\label{sec:related_works}
    \textbf{Dynamic Whole-Body Loco-Manipulation.} Existing dynamic loco-manipulation methods can be broadly categorized into model-based WBC, hierarchical learning-based control, and unified RL policies. Model-based approaches such as whole-body model predictive control (MPC) explicitly optimize robot dynamics, contact constraints, and task-space objectives, enabling precise EE control and physically consistent interaction, but often require accurate models of contact and external forces~\citep{sleiman2021unified, chiu2022collision, rigo2024hierarchical, molnar2026whole}. Hybrid methods such as RAMBO~\citep{cheng2025rambo} combine model-based WBC with RL to improve robustness to unmodeled dynamics while retaining the precision of online optimization. Among learning-based methods, Deep Whole-Body Control~\citep{fu2022deep} pioneered the use of a unified RL policy for coordinated loco-manipulation, while hierarchical learning frameworks, such as ~\citep{liu2024visual, liu2026mlm, jung2026dynamic, ma2025learning}, achieve increasingly diverse capabilities through high-level command or trajectory generation and low-level whole-body tracking. More recent methods explicitly consider EE force interactions: FALCON~\citep{zhang2025falcon} trains humanoid policies with a 3D EE force curriculum for implicit force adaptation, while UniFP~\citep{zhi2025learning} learns a policy capable of both position and force control without relying on physical force sensors. Despite these advances, most methods either model forces explicitly, inject force perturbations during training, or infer interaction effects indirectly from proprioception, rather than directly observing the internal grasp state at the contact interface, including pressure distribution, slip tendency, or loss of contact. Our work targets this missing layer of interaction information by incorporating tactile array observations from the gripper into the policy.

    \textbf{Loco-Manipulation with Tactile Sensing.} Recent advances in tactile sensing have enabled legged robots to perceive rich contact cues, including ground interaction, local deformation, contact shear/force, terrain properties, and payload stability~\citep{lin2025locotouch, shi2024foot, song2024tactid, stone2020walking}. For whole-body loco-manipulation policies, tactile sensors have been mostly introduced for contact-rich manipulation, including diffusion-based, transformer-based, and vision-language-action-style frameworks~\citep{11037531, niu2026learning, zhou2026learning}. However, most of these systems perform manipulation in a quasi-static manner, rather than achieving fully synchronized manipulation during locomotion. 
    This challenge becomes more pronounced when the task requires precise grasp regulation, where the robot must coordinate whole-body motion with tactile feedback to maintain stable yet adaptive contact forces. To address this gap, our system leverages tactile grippers as active policy inputs for dynamic object interaction under external disturbances, variable forces, and sudden release events in an end-to-end manner.

    \textbf{Tactile Simulation.} Although vision-based tactile sensors such as the GelSight family~\citep{yuan2017gelsight} provide rich image-based contact information, accurately simulating their measurements remains challenging. Existing rendering-based tactile simulators often struggle to match real-world tactile images, leading to a large sim-to-real gap~\citep{10912733}. In contrast, low-dimensional array-based tactile sensors offer a simpler and more simulation-friendly alternative, where each taxel measures contact intensity primarily correlated with local normal force~\citep{huang3d, huang2025vt}.  This compact representation is easier to simulate and transfers more reliably between simulation and the real world, enabling scalable parallel training of reinforcement learning policies.

\textbf{Comparison with related works.} We compare our work with representative prior works on legged dynamic loco-manipulation to highlight our differences and contributions. A summary of key feature differences is provided in Tab.~\ref{tab:comparison}. TAC-LOCO differs from prior works by combining unified whole-body control with explicit online grasp adaptation for dynamic loco-manipulation. Particularly, in contrast to LocoTouch~\citep{lin2025locotouch}, which converts distributed tactile readings into a binary contact map to indicate contact presence, our method preserves the continuous spatial distribution of normalized force readings from the tactile arrays.

\begin{table}[hbt]
    \centering
    \small
    \caption{Comparison of existing works and ours.}
    \label{tab:comparison}
    \begin{tabular}{l|cccc}
        \toprule
        \textbf{Method} & \textbf{Unified Policy} & \textbf{Contact sensing} & \textbf{Tactile-informed} & \textbf{Online grasp regulation} \\
        \midrule
        Deep WBC~\citep{fu2022deep} & Yes & No & No & No \\
        VBC~\citep{liu2024visual} & No & No & No & No \\
        FALCON~\citep{zhang2025falcon} & No & Implicit; external & No & No \\
        UniFP~\citep{zhi2025learning} & Yes & Implicit; external & No & No \\
        LocoTouch~\citep{lin2025locotouch} & - & Direct; external & Binary & - \\
        \midrule
        \textbf{TAC-LOCO (Ours)} & Yes & Direct; in-hand & Normalized & Adaptive \\
        \bottomrule
    \end{tabular}
\end{table}

\section{Problem Definition}
\label{sec:problem}
    We formulate the problem as tactile-informed dynamic loco-manipulation of an already-grasped object. Object search, navigation, grasp selection, and reaching-to-grasp are not considered in this work, as they do not involve tactile information and can be can be handled by upstream perception and manipulation modules~\citep{fu2022deep, liu2024visual}. We isolate the post-grasp control problem, when the robot is assumed to have established initial gripper-object contact at the object's initial position $\mathbf{p}_0^{\mathcal{W}}$ before the task begins. Given a desired EE trajectory and a non-trivial locomotion velocity command, the robot must coordinate base, leg, and arm movements while pulling the object. 

    The grasped object is subject to an unknown time-varying external force $\mathbf{F}_{\mathrm{ext},t}$ that may vary smoothly or change suddenly. These interactions can alter the internal grasp condition and may induce slip or contact loss. The control objective is to reach the desired final EE position $\mathbf{p}^{\mathcal{B}}_{f}$ in the robot body frame without losing object contact, while maintaining the desired body velocity under stable locomotion. We define success as completing the target EE motion without losing the grasp. This problem setting isolates the challenge of using tactile feedback for grasp-aware whole-body control under uncertain external forces. 

\section{Methodology}
\label{sec:method}

\begin{figure}[!t]
    \centering
     \includegraphics[width=1.0\linewidth]{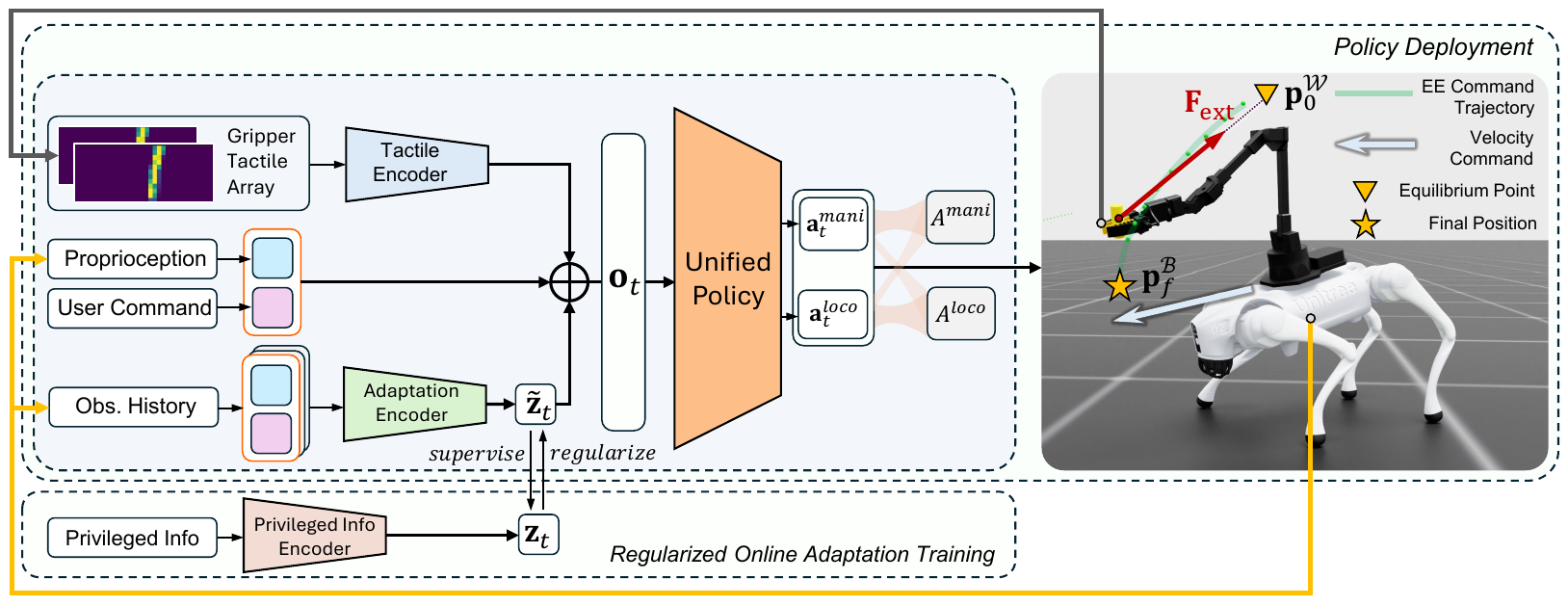}
    \caption{Framework of TAC-LOCO. We encode high-dimensional tactile array information into a lower-dimensional latent representation, then include it in the observation of the unified loco-manipulation policy. The policy is deployed zero-shot and achieves grasp stability under time-varying external forces during dynamic loco-manipulation.}
    \label{fig:framework}
\end{figure}

    \textbf{Overview.} We propose TAC-LOCO, a tactile-augmented unified WBC framework for dynamic loco-manipulation. The policy extends the framework of~\citep{fu2022deep}, where a single multilayer perceptron (MLP) jointly controls the manipulator and leg joints instead of relying on separate locomotion and manipulation controllers, dynamically optimizing whole-body actions to address inherently coupled upper and lower-body motion. Our framework further incorporates tactile feedback from the gripper to inform the policy about the internal grasp state during dynamic interaction, since slip, impending object contact loss, and excessive grasping force are difficult to infer from proprioception alone. We address this by encoding tactile-array observations into a compact latent representation and designing an MDP that enables online grasp regulation during dynamic loco-manipulation. An overview of our framework is shown in Fig. \ref{fig:framework}.

    \textbf{Policy architecture.} During training, the grasped object is subjected to randomized time-varying external forces, including both gradual load changes and sudden release events. This setting follows our problem formulation and reflects deployment scenarios in which an object is pulled against an uncertain load that may change smoothly or vanish abruptly, such as when a fruit stem snaps during pulling. The actor observes only proprioception, action history, and tactile measurements, restricting the policy to observations available during deployment. For each policy inference, tactile array signals from the two gripper fingers are encoded into a compact tactile latent with a two-layer MLP. This compression keeps the tactile representation comparable in dimension to the other observation components, avoiding direct concatenation of high-dimensional tactile arrays that could dominate the policy input. The tactile latent is concatenated with robot proprioception and an encoded history representation, and then passed to the actor network as $\mathbf{O}_t$. The actor uses a shared backbone to extract common whole-body features and branches into separate leg and arm action heads. The leg action head $\mathbf{a}_t^{\mathrm{loco}}$ outputs desired leg joint positions, while the arm action head $\mathbf{a}_t^{\mathrm{mani}}$ outputs desired arm and gripper joint positions. Although actions are partitioned into leg and manipulation heads for credit assignment, both heads share a common observation backbone and are optimized jointly, yielding a single unified policy at deployment. These desired joint positions are sent to low-level motor or servo controllers and tracked through joint-level PD control. 

    \textbf{Policy training.} The policy is trained with rewards grouped into locomotion and manipulation categories. Locomotion rewards $r_{loco}$ encourage commanded base motion, stability, and efficient leg behavior, while manipulation rewards $r_{manip}$ encourage EE tracking and stable object interaction. 
    Inspired by the advantage-mixing idea in~\citep{fu2022deep}, the locomotion and manipulation advantage functions, $A_{loco}$ and $A_{manip}$, are first used to primarily update the corresponding leg and arm action heads, respectively, reducing the difficulty of credit assignment in the high-dimensional action space. A scheduled advantage-mixing mechanism then gradually couples the two objectives, allowing the policy to learn coordinated whole-body behaviors in which locomotion supports tactile-informed manipulation to achieve grasp stability under external disturbances. We train the tactile-informed WBC policy using proximal policy optimization (PPO)~\citep{schulman2017proximal}, a widely used and effective on-policy RL algorithm. 
    The observation space, reward terms, and PPO hyperparameters are reported in the Appendix. 

\subsection{Simulation Setup}
\label{sec:sim}
    \textbf{Task Simulation.} We implement our framework in Isaac Lab~\citep{mittal2025isaaclab} using a Unitree Go2 quadruped robot with an Interbotix WidowX 250 arm on its back. At the beginning of each episode, the arm is initialized at a randomized configuration, and the object is placed at the center of the gripper with the gripper width set close to the object diameter, simulating an already-established grasp condition. To model unknown external forces on the object, we use a spring-damper force model whose equilibrium position is set to $\mathbf{p}_0^{\mathcal{W}}$, as illustrated in Fig. \ref{fig:framework}. The stiffness $k$, damping $c$, and force-release threshold $F_\mathrm{thr}$ are randomized across episodes to cover a range of realistic deployment conditions. When the external force exceeds the sampled threshold, the force is removed for the remainder of the episode, modeling sudden release events such as detachment when pulling an object from a resisting source. Here the external force $\mathbf{F}_{\mathrm{ext},t}$ is defined as:
    \begin{equation}
    \mathbf{F}_{\mathrm{ext},t} = 
    \begin{cases}
    \mathbf{0}, 
    & \exists\, T < t \;\text{such that}\; 
    \|\mathbf{F}_{\mathrm{ext},T}\| > F_\mathrm{thr}, \\[2mm]
    k\mathbf{p}_{e,t} + c(\dot{\mathbf{p}}^{\mathcal{W}} \cdot \hat{\mathbf{p}}_{e,t})\hat{\mathbf{p}}_{e,t}, 
    & \text{otherwise}.
    \end{cases}
    \end{equation}
    where $\mathbf{p}_t^\mathcal{W}$ is the object position, $\mathbf{p}_{e,t} = \mathbf{p}_0^{\mathcal{W}}-\mathbf{p}_t^\mathcal{W}$ and $\hat{\mathbf{p}}_{e,t}$ denotes the unit vector in the direction of $\mathbf{p}_{e,t}$. In addition to standard terminations such as robot falling, we terminate the episode \(2.0\,\mathrm{s}\) after complete object loss from the gripper, avoiding training on post-failure states while discouraging abrupt recovery behaviors that would be undesirable for safe deployment.

    \textbf{Tactile Simulation.} Following prior work~\citep{huang2025vt}, we simulate tactile contacts in Isaac Lab using the TacSL formulation~\citep{10912733}: for each taxel point $\mathbf{x}_i$, contact is detected by querying the signed distance field (SDF) of the rigid object. Negative SDF values indicate penetration, with depth $d_i=\max(-\phi(\mathbf{x}_i),0)$ after correcting for non-uniform mesh scaling. The contact normal $\mathbf{n}_i$ is obtained from the normalized SDF gradient and transformed back to the world frame. The elastomer-object interaction is modeled as compliant contact with PhysX stiffness and damping parameters, equivalent to a Kelvin--Voigt spring-damper model. Small collision offsets allow controlled interpenetration, improving contact stability and producing smoother tactile signals. For each taxel, the normal force is computed as $\mathbf{f}_{n,i}=k_n d_i \mathbf{n}_i$, transformed into the local tactile frame, projected onto the tactile surface normal, and flattened into a $12\times32$ normal-force observation for each sensor.
    
    To reduce the sim-to-real gap while preserving spatial contact patterns, we normalize each simulated tactile frame independently. Given the raw force map $\mathbf{F}\in\mathbb{R}^{12\times32}$, we set $F_{ij}=|f_{n,ij}|$ and $F_{\max}=\max_{i,j}F_{ij}$. The normalized value is $\tilde{F}_{ij}=F_{ij}/F_{\mathrm{ref}}$ when $F_{\max}<\tau_F$, and $\tilde{F}_{ij}=F_{ij}/F_{\max}$ otherwise, where $\tau_F$ is a near-zero contact threshold and $F_{\mathrm{ref}}$ is a fixed reference scale. The resulting map is clipped to $[0,1]$ and flattened as the tactile observation. Additional implementation details, sim-to-real comparisons, and tactile-domain randomization are provided in the Appendix.


\subsection{Observation Design}
\label{sec:observation}
The privileged states available to the privileged encoder include not only quantities commonly used for trajectory tracking and dynamics adaptation, such as the base linear velocity and EE position, but also object contact information not directly available on hardware. Specifically, we include the object position and velocity relative to the end effector, which indicate object retention and potential slip, together with the internal contact forces between the object and gripper in both normal and shear directions. These privileged quantities provide supervision for evaluating whether the object is securely grasped, sliding within the gripper, or approaching contact loss. During deployment,
 the actor must infer the internal grasp state from proprioception and tactile array measurements. This design encourages the policy to use tactile feedback as the deployable sensing modality for grasp-aware whole-body control under external disturbances. 
 A detailed list of observations for each module of our work is listed in the Appendix.

\subsection{Grasp Stability Reward Design}
\label{sec:reward}
The grasp stability reward is a subset of the manipulation reward, designed to regulate the internal object-gripper interaction during dynamic loco-manipulation. In addition to tracking the desired EE trajectory, the policy must maintain a secure but not overly tight grasp. We therefore decompose the grasp stability reward into three terms: $ r_{\mathrm{grasp}} = r_{\mathrm{force}} + r_{\mathrm{slip}} + r_{\mathrm{retention}}$.
The force moderation term $r_{\mathrm{force}}$ penalizes excessive internal grasping force between the gripper and the object, encouraging compliant contact rather than over-constraining the object. The slip term $r_{\mathrm{slip}}$ penalizes relative velocity between the object and the EE, discouraging incipient sliding within the grasp. The retention term $r_{\mathrm{retention}}$ penalizes excessive relative displacement between the object and the gripper, preventing complete loss of contact or object dropping. 
In summary, these terms encourage the policy to use tactile feedback to maintain stable object retention while avoiding unnecessarily large grasping forces. A complete reward table is provided in the Appendix. 


\section{Simulation Results}
\label{sec:result}
\begin{wrapfigure}[16]{r}{0.55\textwidth}
    \vspace{-0.7cm}
    \centering
    \includegraphics[width=\linewidth]{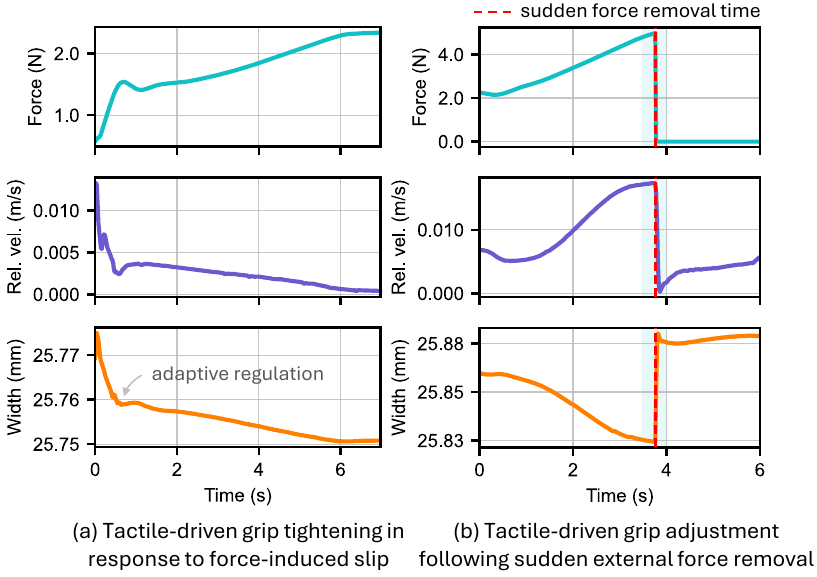}
    \caption{Grasping width changes during object-EE relative velocity and time-varying external force.}
    \label{fig:width}
\end{wrapfigure}
We first evaluate TAC-LOCO in simulation to study how tactile feedback affects grasp regulation during dynamic loco-manipulation. We design experiments to answer the following questions: \textbf{(1)} Can the learned policy adapt the gripper width in response to changing external forces and EE motion (Sec. \ref{sec:adaptation})? \textbf{(2)} Does tactile feedback improve task success and reduce object slip (Sec. \ref{sec:loco-manipulation})? \textbf{(3)} Can the learned policy track long EE trajectories and body commands while maintaining object retention (Sec. \ref{sec:additional_sim})? \textbf{(4)} How does each component of the grasp stability reward contribute to force moderation, slip suppression, and object retention (Sec. \ref{sec:ablation_reward})?

Due to the dynamic nature of our tasks, readers are strongly encouraged to view the supplementary video for a better visualization of our results.

\subsection{Grasping Width Adaptation}
\label{sec:adaptation}

\textbf{Tactile feedback enables adaptive grasp regulation under time-varying external forces.} We examine whether the learned policy uses tactile feedback to regulate grip strength online rather than maintaining a fixed conservative grasp. Fig.~\ref{fig:width} shows two representative simulation rollouts, plotting the external force applied to the object $\mathbf{F}_{\mathrm{ext},t}$, the relative velocity between the object and EE, and the gripper width over time. In the left subplot, the increasing external force induces object slip, indicated by non-zero relative velocity. The policy responds by rapidly reducing the gripper width and further tightening the grasp as the force continues to grow, until the relative velocity is suppressed. In the right subplot, the policy similarly tightens the grasp under increasing force. However, when the force suddenly disappears, the gripper width increases again, relaxing the grasp. These results show that the policy does not rely on overly firm grasping, but adaptively regulates grip online based on tactile cues from object motion and external disturbances.

\subsection{No-slip Object Loco-manipulation with Tactile Feedback}
\label{sec:loco-manipulation}

\begin{wrapfigure}[12]{r}{0.5\textwidth}
    \vspace{-0.45cm}
    \centering
     \includegraphics[width=\linewidth]{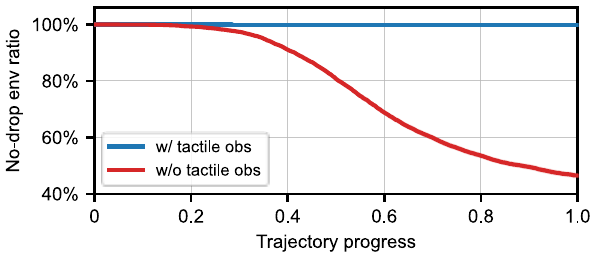}
    \caption{Fraction of rollouts maintaining grasp contact with the object over trajectory progress. Tactile-informed policy substantially improves slip-free completion of the trajectory.}
    \label{fig:retention}
\end{wrapfigure}

\textbf{Tactile feedback improves robustness under dynamic loco-manipulation.} We next evaluate whether tactile feedback improves task success and reduces object slip during dynamic loco-manipulation. We compare our method, which uses tactile arrays on the gripper fingers, against a proprioception-only baseline that does not receive tactile observations but still observes the gripper width through joint position feedback. For both methods, we sample $8000$ randomized combinations of body-velocity commands and EE trajectory commands and roll out the policy under randomized external forces. When the object is dropped, we record the trajectory progress at failure, defined as the current time divided by the total trajectory duration. 

 Figure \ref{fig:retention} plots the fraction of rollouts that remain grasping the object over the trajectory progress. As the trajectory progresses, the proprioception-only baseline exhibits an increasing object-drop rate due to the magnitude and variance of the external force growing as the object is pulled farther from its equilibrium. As a result, only $46.4\%$ of the rollouts without tactile observations complete the trajectory successfully. In contrast, our tactile-informed policy maintains object retention for the entire trajectory $99.9\%$ of the time. This indicates that tactile observations provide critical information about the contact state, enabling the policy to adjust the gripper and whole-body motion before slip develops into complete loss of contact. 

\subsection{Additional Simulation Evaluation}
\label{sec:additional_sim}
\begin{figure}[!h]
\centering
\includegraphics[width=1\textwidth]{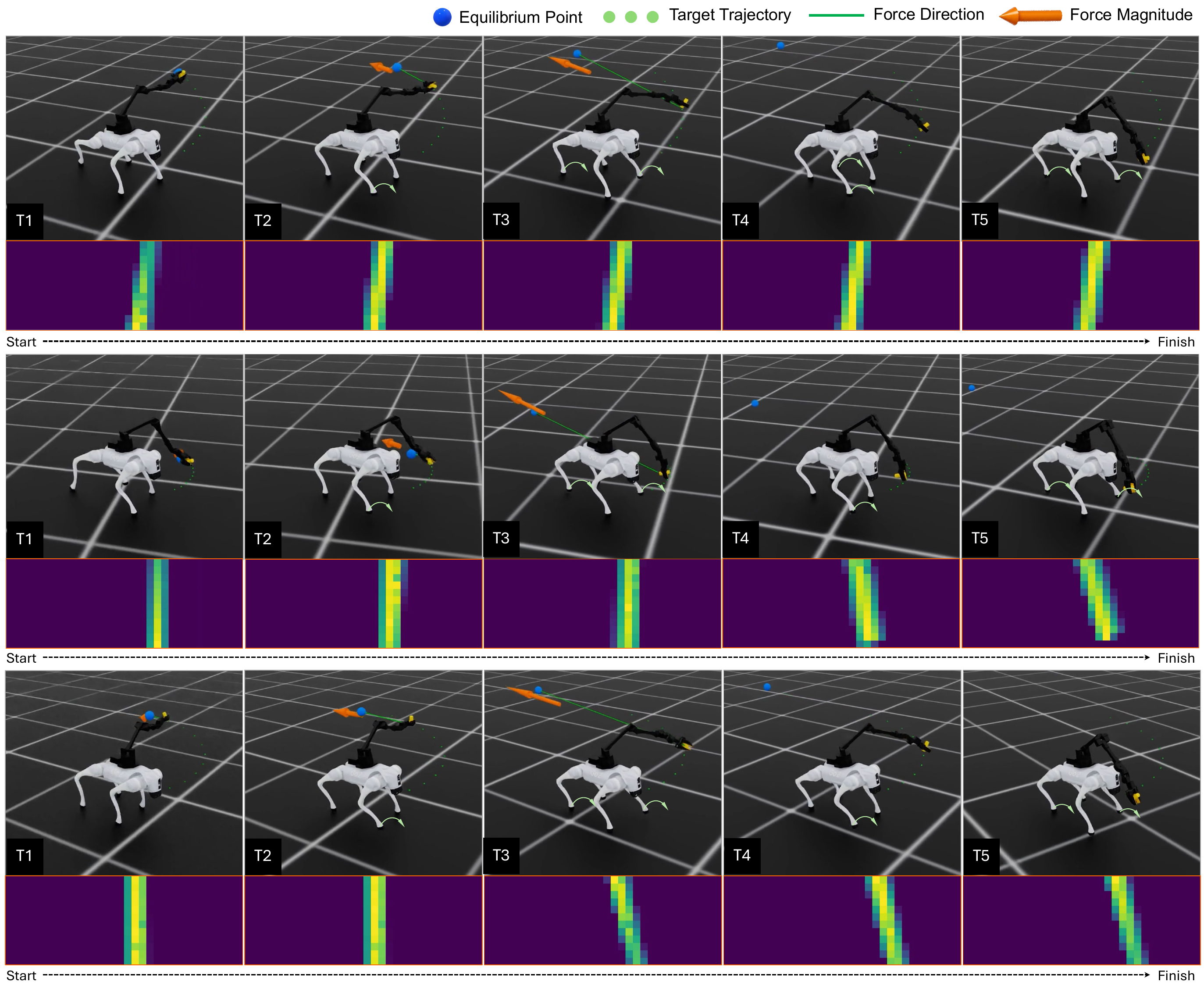}
\caption{Simulation evaluation under time-varying external forces with a wider workspace. Each row shows a representative robot trajectory together with the tactile observation from one gripper finger. From $T_1$ to $T_3$, the external force pulling the object toward the equilibrium point gradually increases during loco-manipulation and is suddenly released at $T_4$. The robot maintains balance and grasp stability while reaching the target position at $T_5$.}
\label{fig:sim_workspace}
\end{figure}

\textbf{The policy maintains grasp stability over large workspace motions.} To further characterize the loco-manipulation capability of the proposed method, we visualize simulation rollouts  with a large workspace and long EE trajectories. As shown in Fig.~\ref{fig:sim_workspace}, the robot moves the grasped object from an initial position far to the side of the body to a final target position in front of the robot, requiring coordinated locomotion, arm motion, and gripper regulation over a large spatial displacement. This result illustrates that the learned policy can track long EE trajectories while maintaining object retention under time-varying external forces. The supplementary video further visualizes the change in the external force applied to the object, together with the corresponding tactile array observations throughout the trajectory.

\begin{figure}[!h]
\centering
\includegraphics[width=1\textwidth]{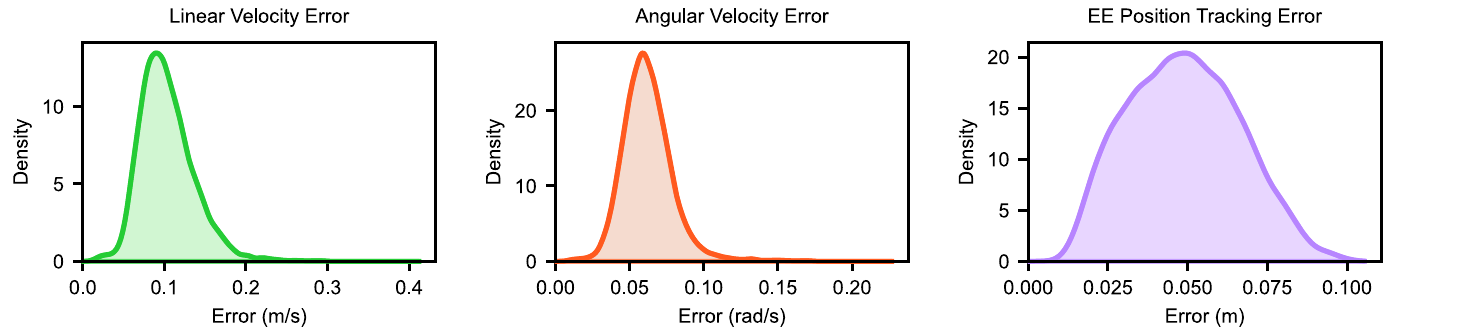}
\caption{Distributions of the linear-velocity, angular-velocity, and end-effector position tracking errors over $8000$ trajectories in simulation.}
\vspace{1.5cm}
\label{fig:metrics}
\end{figure}

\textbf{The policy preserves whole-body command tracking during tactile-informed grasp regulation.} We further evaluate command tracking performance across randomized simulation rollouts. Specifically, we measure the tracking errors for body linear velocity, body yaw rate, and end-effector position along the commanded trajectory. Fig.~\ref{fig:metrics} reports the distribution of the average tracking error for each metric over $8000$ rollouts, where each rollout samples a randomized end-effector trajectory and body velocity command. The tracking errors remain low across randomized rollouts, demonstrating accurate command tracking for both locomotion and end-effector motion comparable to that reported in baseline work~\citep{fu2022deep}. 

\subsection{Ablations on Reward Design}
\label{sec:ablation_reward}

\begin{wrapfigure}[18]{r}{0.48\textwidth}
    \vspace{-1.35cm}
    \centering

    \includegraphics[width=\linewidth]{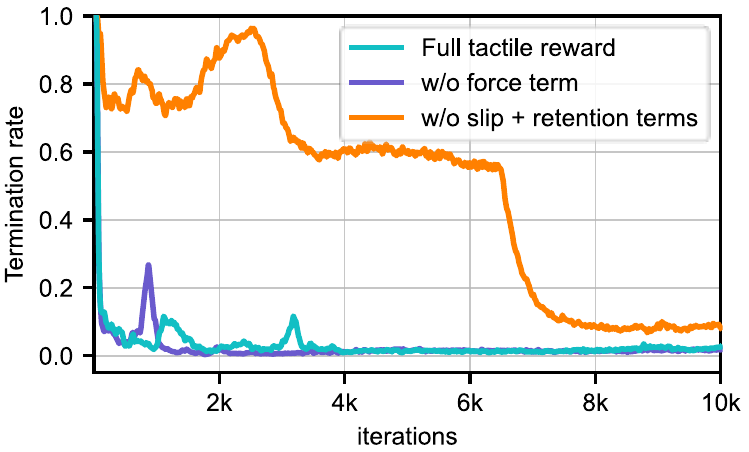}
    \caption{Ratio of environments terminated over policy update iterations. Slip and retention rewards improve task success.}
    \label{fig:ablation_mdp}

    \vspace{0.2cm}

    \resizebox{\linewidth}{!}{
    \begin{tabular}{lcc}
        \toprule
        Policy Variant 
        & Avg. Force (N) $\downarrow$
        & Width (mm) $\uparrow$\\
        \midrule
        TAC-LOCO full reward 
        & 13.11 
        & 24.65 \\
        w/o Force Reg. Reward 
        & 24.76 
        & 22.20 \\
        w/o Retention/Slip Reward 
        & 25.74 
        & 21.62 \\
        \bottomrule
    \end{tabular}}
    \captionof{table}{\small Ablation study on average tactile force and gripping width. Force is averaged over the left and right elastomer contact force magnitudes.}
    \label{tab:ablation_force_width}

\end{wrapfigure}

\textbf{Grasp stability rewards promote robust yet force-efficient object retention.} We examine how the grasp stability reward components contribute to force moderation, slip suppression, and object retention. We first study the slip and retention terms, which provide dense guidance for keeping the object within the gripper. As shown in Fig. \ref{fig:ablation_mdp}, removing these terms forces the policy to rely mainly on episode termination from object drops. This sparse, binary signal makes credit assignment difficult, as it does not reveal whether failure results from gradual slip, excessive EE motion, or complete contact loss.

We further run an ablation on the force moderation term to evaluate whether it prevents overly conservative grasping. Removing this term does not impact the task success rate; however, this results in substantially higher internal grasping force throughout the episode (Tab. \ref{tab:ablation_force_width}). Compared with the ablation without force moderation, our full reward design uses only $53\%$ of the average grasping force while maintaining comparable object retention performance. This indicates that, without explicit force moderation, the policy tends to maximize task success through conservative grasping, whereas the proposed reward encourages a more adaptive grasp that is sufficient for slip prevention without applying unnecessary contact force.

\section{Deployment Results}
\label{sec:deploy}

    We deploy the TAC-LOCO policy zero-shot on hardware, using normalized tactile observations and tactile domain randomization to reduce the sim-to-real gap. For the robot hardware, two \emph{FlexiTac} tactile sensors \citep{huang2025vt} are mounted on fin-shaped parallel TPU fingers, providing the same sensing resolution ($12 \times 32$) and contact area ($25 \times 66\,\text{mm}^2$) as in simulation. To reproduce the force-changing task, we use springs with stiffness values of $6.1$ and $12.3$ ${\mathrm{N/m}}$ and magnets with breaking forces ranging from $2$ to $5\,\mathrm{N}$, matching the randomized force profiles used in simulation. In hardware, the grasped object is a cylindrical object with the same shape and size as the object used during training, connected to a spring-magnet assembly, where the opposite magnet is fixed to a rigid cable. As the robot pulls the object, the spring force increases until it exceeds the magnetic breaking force, causing magnet separation and producing a sudden force-release event. The experiment setup is depicted in Fig. \ref{fig:deploy}. Each trial starts from an already-established grasp, using the same initial gripper configuration as in simulation and producing clear tactile contact features. In the real-world experiments, the commanded body velocity and EE trajectory are selected from predefined command sets grouped into two difficulty levels, and $10$ trials are conducted for each setting. Detailed trajectory parameters are specified in the Appendix.

\subsection{Whole-body Motion and Dynamic Loco-manipulation}
\label{sec:loco_manipulation}

\begin{wraptable}[5]{r}{0.5\textwidth}
    \vspace{-0.5cm}
    \centering
    \resizebox{\linewidth}{!}{
    \begin{tabular}{lcc}
        \toprule
        Task 
        & Large EE Motions
        & Loco-manipulation\\
        \midrule
        Success Rate 
        & 1.0 
        & 0.9 \\
        \bottomrule
    \end{tabular}}
    \caption{Success rates of policy deployment of two difficulty levels.}
    \label{tab:success_rate}
    \vspace{-0.4cm}
\end{wraptable}

\begin{figure}[!t]
    \centering
    \includegraphics[width=1.0\linewidth]{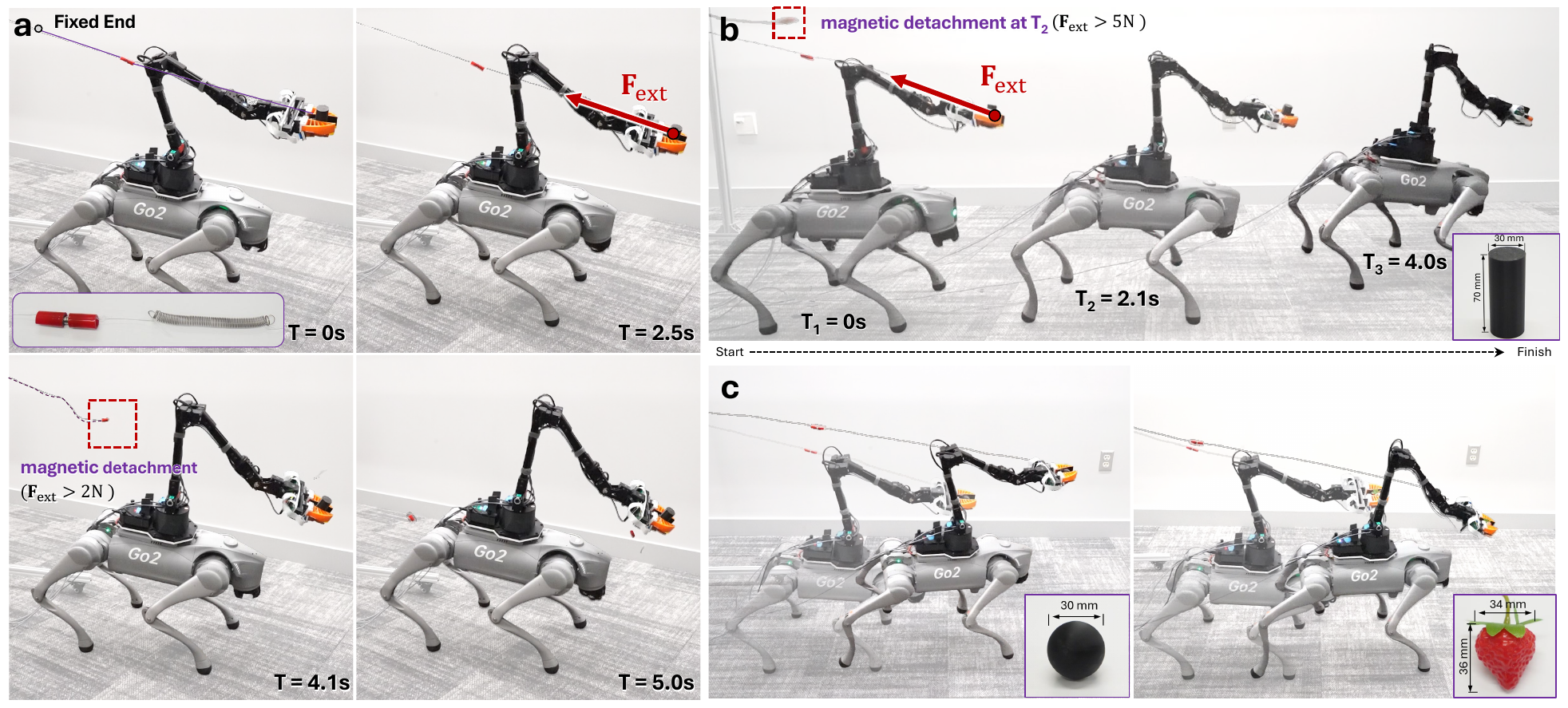}
    \caption{Real-world evaluation tasks. 
    (a) Manipulation under time-varying external forces, where magnets and springs generate controlled stiffness, external loading, and sudden force-release events. 
    (b) Dynamic loco-manipulation, where the robot tracks base-velocity commands while transporting a grasped cylinder and recovering from sudden force release. 
    (c) The trained policy generalizes to unseen objects during loco-manipulation.}
    \vspace{-0.4cm}
    \label{fig:deploy}
\end{figure}

    In Fig. \ref{fig:deploy} (a), the policy successfully maintains the grasp while executing large EE motions and whole-body posture changes. The same policy also enables dynamic loco-manipulation, where the robot walks across the workspace while pulling the object and maintaining contact throughout the motion, shown in Fig. \ref{fig:deploy} (b). In both tasks, object contact was sustained across sudden changes in the object’s external force caused by magnet separation. Readers may refer to the supplementary video. 

\subsection{Generalizability in Object Shapes and Sizes}
\label{sec:OOD}

    We further evaluate whether the same policy generalizes to object geometries not seen during training. Without any fine-tuning, the policy successfully completes the task while grasping a sphere and an artificial strawberry, demonstrating that it does not rely on a fixed object shape or contact pattern. We attribute this generalization to tactile-domain randomization and the normalized tactile representation, which encourage the policy to react to changes in contact area and contact distribution rather than memorize explicit object geometry.

\subsection{Robustness to disturbance forces on the object} 
\label{sec:OOD_object}
\begin{figure}[!h]
\centering
\includegraphics[width=0.8\textwidth]{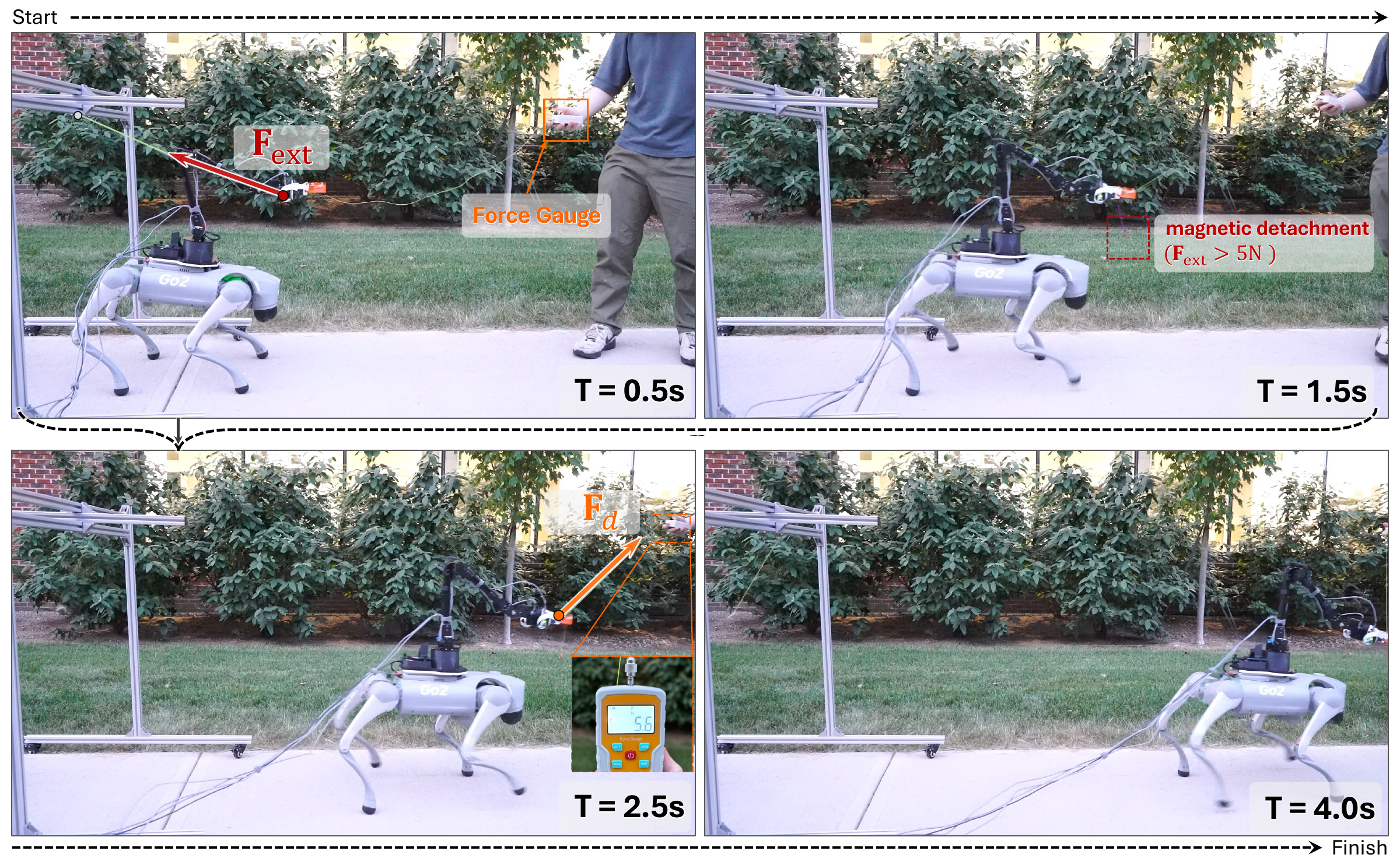}
\caption{Real-world robustness to unmodeled object disturbances. During loco-manipulation, randomized perturbation forces up to $5.6~\mathrm{N}$ are manually applied to the grasped artificial strawberry in directions independent of the nominal spring--damper force. The robot maintains stable locomotion and object retention while completing the task.}
\label{fig:ood}
\end{figure}

We further test the robustness of the deployed policy against additional object disturbances beyond the spring-damper external force modeled in training. During the loco-manipulation task, we manually apply randomized perturbation forces $\mathbf{F}_d$ to the grasped object in directions independent of the nominal spring-damper force, with a maximum magnitude of $5.6$ N, which is comparable in order of magnitude to the modeled external force. Shown in Fig.~\ref{fig:ood}, despite these disturbances, the robot successfully completes the task, maintaining stable locomotion and object retention, demonstrating that the learned policy can tolerate unmodeled external perturbations during real-world dynamic loco-manipulation.

\subsection{Grasp regulation behavior on hardware}
\begin{figure}[!h]
\centering
\includegraphics[width=0.9\textwidth]{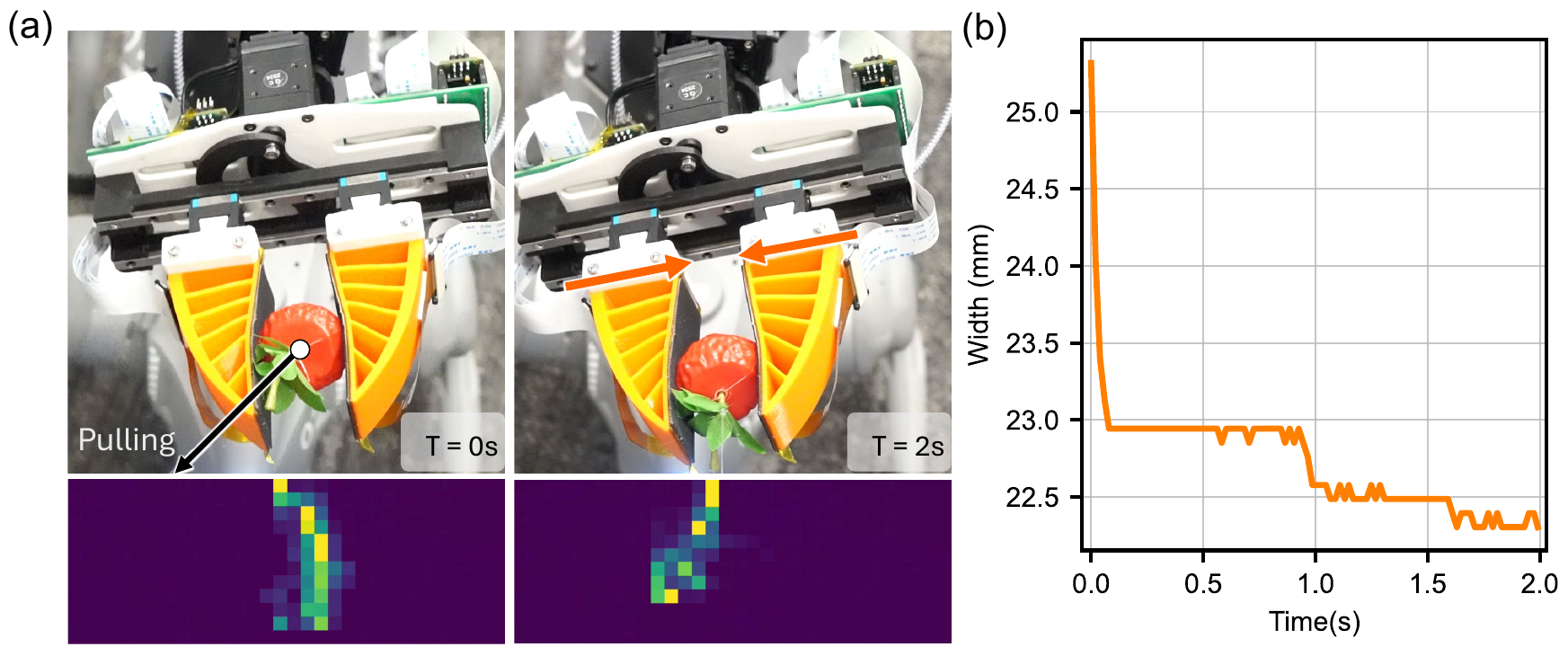}
\caption{Grasp regulation behavior of the trained policy.
(a) The artificial strawberry is pulled from the center toward the edge of the elastomer surface.
(b) The gripper width is adjusted online in response to in-hand slip and changes in the applied force, maintaining a stable grasp while preventing excessive gripping force.}
\label{fig:hardware_width}
\end{figure}

As shown in Fig.~\ref{fig:hardware_width}, we manually pull a previously unseen grasped object, an artificial strawberry, outward from the gripper to induce in-hand slip. Once slip occurs, the policy rapidly decreases the gripper width and continues tightening the grasp as the object approaches the fingertip edge, consistent with the simulation results in Sec.~\ref{sec:adaptation}. Compared with simulation, the hardware response exhibits larger gripper-width adjustments, likely due to differences in fingertip compliance and material properties between the simulated and real systems. Nevertheless, the policy adapts until a stable tactile contact pattern is restored. Similar behavior is observed across other unseen object geometries, suggesting that the policy relies on generalizable contact features, such as the spatial distribution and relative intensity of tactile signals, rather than object-specific tactile patterns.

\subsection{Baseline comparison} 
We compare our method with Deep WBC~\citep{fu2022deep} deployed on hardware, which uses a similar unified policy training framework for coordinated locomotion and manipulation, but does not include tactile observations or actions for adaptive gripper control. During deployment, the gripper width is set to a fixed value matching the initial gripper width used in both training and deployment of our tactile-informed policy. This gripper width firmly grasps the object under no external load, while avoiding an overly conservative grasp. 

\begin{figure}[!h]
\centering
\includegraphics[width=1\textwidth]{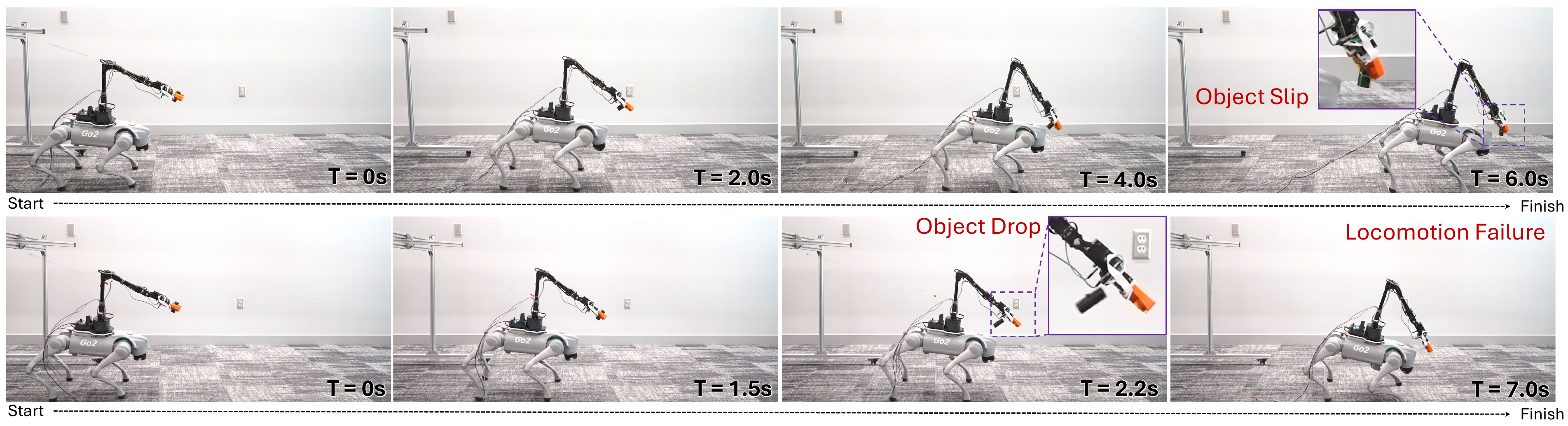}
\caption{Benchmarking against~\citep{fu2022deep}. The baseline uses a fixed gripper width and therefore cannot actively compensate for changes in the external force. The top row shows gradual in-hand object slip during loco-manipulation. The bottom row shows a failure case in which the object is dropped, followed by locomotion failure.}
\label{fig:dwc}
\end{figure}

We conduct 10 hardware trials using the same set of loco-manipulation commands as in Sec.~\ref{sec:deploy}. Table~\ref{tab:baseline} compares the success rates of our policy and the baseline, showing that our tactile-informed policy achieves a substantially higher success rate. Figure~\ref{fig:dwc} shows that the baseline suffers from gradual object slip during the task. Since the gripper width remains fixed, the policy cannot actively suppress slip once the external force changes.

\begin{table}[!ht]
    \centering
    \caption{Deployment success rate comparison with baseline.}
    \label{tab:baseline}
    \begin{tabular}{lcc}
        \toprule
        Method 
        & Deep WBC~\citep{fu2022deep}
        & TAC-LOCO (Ours)\\
        \midrule
        Success Rate 
        & 0.5
        & 0.9 \\
        \bottomrule
    \end{tabular}
\end{table}

We also observe that the baseline can exhibit unstable locomotion when the object is dropped during deployment. In this case, the sudden loss of object contact induces an abrupt change in the force applied to the end effector, causing large oscillatory body motion and poor body-velocity tracking. We attribute this failure mode to the baseline policy not encountering such sudden interaction changes during training, leading to undesirable motion and locomotion failure. 

\section{Limitations}
\label{sec:limitations}
Our method has two main limitations. First, although tactile feedback enables the policy to reduce internal grasping force, the current simulator does not perfectly reproduce the real-world deformation of the gripper and tactile array. In particular, after normalizing tactile readings for sim-to-real transfer, the policy can reliably use tactile signals to infer relative changes in contact state, but not to enforce an exact absolute grasp-force threshold. Second, our hardware uses a servo-controlled arm whose actuator dynamics and control are difficult to accurately identify for simulation. As a result, the deployment results may not fully reflect the potential of the proposed method; we expect that a proprioceptive, backdrivable torque-controlled arm would enable stronger dynamic performance.


\section{Conclusion}
\label{sec:conclusion}
In this work, we presented TAC-LOCO, a tactile-informed unified reinforcement learning framework for quadrupedal loco-manipulation under unknown and time-varying external forces. By integrating compact tactile representations with proprioceptive observations, the proposed policy jointly coordinates locomotion, arm motion, and adaptive gripper control. Simulation and zero-shot hardware experiments demonstrate that tactile feedback improves object retention, reduces unnecessary grasping force, and enables robust dynamic manipulation across varying interactions and object geometries. These results highlight the potential of tactile sensing for reliable whole-body control.


\clearpage


\bibliography{bibliography}  

\clearpage
\appendix

\section{Details for Tactile Simulation Implementation}
\subsection{Tactile Signal Computation}
\label{sec:tac_signal}
\textbf{Taxel Generation:}
Each tactile sensor is represented by a regular $12 \times 32$ taxel grid, matching the physical sensor layout with a center-to-center spacing of $2~\mathrm{mm}$ between adjacent taxels. The taxel locations are generated by projecting this planar grid onto the elastomer mesh through ray casting. Each ray is cast along the thin axis of the elastomer toward the contact surface, and its first intersection with the mesh defines the corresponding taxel location.

\textbf{Contact Modeling Parameters:}
Following TacSL~\citep{10912733}, we model the elastomer--object interaction using a Kelvin--Voigt contact model. For each taxel $i$, the normal-force magnitude is computed as
$f_{n,i}=-(k_n d_i+k_d \dot{d}_i)$,
where $d_i$ is the penetration depth and $\dot{d}_i$ is the relative normal velocity. The stiffness coefficient $k_n$ determines the elastic response to penetration, whereas the damping coefficient $k_d$ suppresses contact oscillations induced by rapid changes in penetration depth. In our implementation, we followed TacSL and set $k_n=1.0$ and $k_d=3 \times 10^{-3}$. In addition to the force model, PhysX introduces contact-level parameters, including collision offsets and the elastomer material's compliant-contact stiffness, which determine the realized object--elastomer configuration and thus the taxel penetration depths $d_i$. As detailed in App.~\ref{sec:tac_sim2real}, we randomize these parameters during training to reduce the sim-to-real gap.

\textbf{Tactile Encoder:} 
The normalized tactile observations from the two fingers are flattened and concatenated into a $768$-dimensional vector. This vector is passed through a shared MLP encoder, and the resulting latent representation is concatenated with the proprioception features. The tactile encoder consists of two fully connected layers with ELU activations, mapping the input to a compact $32$-dimensional latent representation through a $128$-dimensional hidden layer. Its parameters are jointly optimized with the actor--critic networks during PPO training.

\subsection{Tactile Sim-to-Real}
\label{sec:tac_sim2real}
\textbf{Normalization:}
Piezoresistive tactile arrays exhibit nonlinear force responses~\citep{huang3d}. Moreover, accurately reproducing soft-contact mechanics in simulation remains challenging. Therefore, rather than calibrating a taxel-wise force--response curve between simulation and the real sensor, we use the normalized tactile signals introduced in Sec.~\ref{sec:sim}.

\begin{figure}[!h]
\centering
\includegraphics[width=1\textwidth]{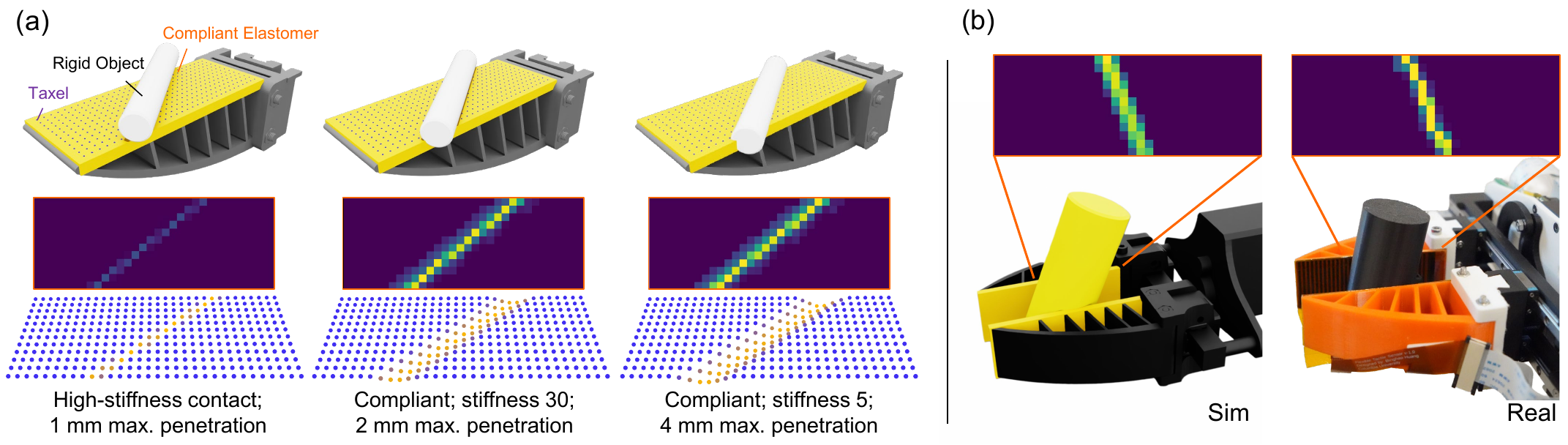}
\caption{Tactile simulation settings and sim-to-real comparison.
(a) Effects of the elastomer stiffness and PhysX penetration-offset parameters on the simulated tactile-array signals.
(b) Comparison of tactile patterns obtained in Isaac Lab simulation (left) and real-world experiments (right) while grasping a 30-mm-diameter cylinder at the same gripper width.}
\label{fig:tac_sim2real}
\end{figure}

\textbf{Contact Modeling Domain Randomization:}
Unlike manipulation tasks with a fixed gripper width~\citep{huang2025vt}, our dynamic loco-manipulation task requires rapid online grasp regulation. To improve robustness, we randomize the elastomer material's compliant-contact stiffness and the PhysX collision offsets, which affect the realized object--elastomer configuration and the resulting taxel penetration depths. As shown in Fig.~\ref{fig:tac_sim2real}(a), these parameters substantially influence the simulated tactile patterns. To reduce the sim-to-real gap caused by imperfect modeling of the compliant fingertips, we apply domain randomization to the compliant-contact stiffness over $[5, 100]$ and to the collision-offset-induced penetration allowance over $[1, 3]~\mathrm{mm}$. We set the tactile simulation frequency to $100~\mathrm{Hz}$ to match the sampling rate of the real hardware. A comparison of the tactile patterns obtained in simulation and real-world experiments is shown in Fig.~\ref{fig:tac_sim2real}(b). The real-world evaluations presented in Fig.~\ref{fig:deploy} and Sec.~\ref{sec:deploy} further demonstrate successful sim-to-real transfer and robust generalization to unseen objects.

\section{Training Details}
\label{sec:training}

We train the proposed policy with PPO~\citep{schulman2017proximal} in Isaac Lab~\citep{mittal2025isaaclab} using $4096$ parallel environments. To facilitate reproducibility, we report the key implementation details of the training setup in this section. These include the randomized environment parameters, command ranges, observation definitions, reward terms, and PPO hyperparameters, which are summarized in Tabs.~\ref{tab:env_cfg}--\ref{tab:ppo_cfg}.


\begin{table}[h]
    \centering
    \caption{\textbf{Environment parameters} for policy training.}
    \label{tab:env_cfg}
    \begin{tabular}{llc}
        \toprule
        \textbf{Parameter} & \textbf{Range / Value} \\
        \midrule
        Ground static friction & $[0.5,\ 1.5]$ \\
        Ground dynamic friction & $[0.4,\ 1.0]$ \\
        Gripper static friction & $[0.4,\ 1.2]$ \\
        Gripper dynamic friction & $[0.3,\ 1.0]$ \\
        Body mass variation & $[-0.5,\ 1.5]$ kg \\
        EE mass variation & $[0.0,\ 0.5]$ kg \\
        Leg motor PD gain multiplier & $[0.9,\ 1.1]$ \\
        Arm servo PD gain multiplier & $[0.8,\ 1.2]$ \\
        Object external force stiffness & $[0.0,\ 18.0]$ N/m \\
        Object external force damping & $[0.1,\ 0.5]$ N/ms$^{-1}$ \\
        Object external force threshold & $[0.5,\ 10.0]$ N \\
        \bottomrule
    \end{tabular}
\end{table}

\vspace{3.0cm}

\begin{table}[h]
    \centering
    \caption{\textbf{Command ranges} during policy training.}
    \label{tab:cmd_cfg}
    \begin{tabular}{llc}
        \toprule
        \textbf{Command} & \textbf{Range} \\
        \midrule
        Body linear velocity $v_{\mathrm{cmd}}$ & $[0.1,\ 0.8]$ m/s \\
        Body yaw rate $\omega_\mathrm{cmd}$ & $[-0.5,\ 0.5]$ rad/s \\
        EE final position $\mathbf{p}^{\mathcal{B}}_{f}$ & $([0.28, 0.51],\ [-0.35,\ 0.35],\ [-0.07,\ 0.20])$ m \\
        EE trajectory total time $t_\mathrm{traj}$ & $[4,\ 8]$ s \\
        \bottomrule
    \end{tabular}
\end{table}

\begin{table}[h]
    \centering
    \caption{\textbf{Policy observations} of the proposed framework.}
    \label{tab:obs_cfg}
    \begin{tabular}{lc}
        \toprule
        \textbf{Observation term} & \textbf{Dimension} \\
        \midrule
        \textbf{Proprioception}\\
        \midrule
        Body angular velocity & $3$\\
        Joint positions & $19$\\
        Joint velocities & $19$\\
        Projected gravity vector & $3$\\
        Previous action & $19$\\
        \midrule
        \textbf{User command}\\
        \midrule
        Body velocity command & $2$\\
        EE position command & $3$\\
        \midrule
        \textbf{Privileged information}\\
        \midrule
        Body mass & $1$\\
        EE mass & $1$\\
        Body linear velocity & $3$\\
        Joint torques/force & $19$\\
        Feet contact & $4$\\
        EE position & $3$\\
        EE orientation & $4$\\
        EE linear velocity & $3$\\
        EE angular velocity & $3$\\
        Object-EE relative position & $3$\\
        Object-EE relative velocity & $3$\\
        Grasping forces & $6$\\
        Proprioception observations & $63$\\
        \bottomrule
    \end{tabular}
\end{table}

\begin{longtable}{ll}
    \caption{\textbf{Notation} used in the reward definitions.}
    \label{tab:notation}\\
    \toprule
    \textbf{Notation} & \textbf{Definition} \\
    \midrule
    \endfirsthead

    \toprule
    \textbf{Notation} & \textbf{Definition} \\
    \midrule
    \endhead

    $\mathbf{p}_\mathrm{obj}$ 
    & Object position \\
    $\mathbf{p}_\mathrm{ee}$ 
    & EE position \\
    $\mathbf{p}_{\mathrm{ee},\mathrm{cmd}}$ 
    & EE position command\\
    $\mathbf{v}_\mathrm{obj}$ 
    & Object linear velocity\\
    $\mathbf{v}_\mathrm{ee}$ 
    & EE linear velocity\\
    $d=\lVert \mathbf{p}_\mathrm{obj}-\mathbf{p}_\mathrm{ee}\rVert_2$ 
    & Object-EE distance \\
    $\mathcal{B}_g$ 
    & Set of gripper-object contact bodies\\
    $\mathbf{F}_{b}$ 
    & Contact force on body $b$ \\
    $v_{x,b}$ 
    & Base linear velocity \\
    $v_{z,b}$ 
    & Base vertical velocity \\
    $\omega_{z,b}$ 
    & Base yaw rate\\
    $\omega_{x,b},\omega_{y,b}$ 
    & Base roll and pitch angular velocities\\
    $\mathcal{J}_{all}$ 
    & Set of all robot leg joints\\
    $\mathcal{J}_{hip}$ 
    & Set of hip joints \\
    $\mathcal{J}_{tc}$ 
    & Set of thigh and calf joints \\
    $c_i$ 
    & Contact indicator for foot $i$ \\
    $t_i$ 
    & Current air-time or contact-time variable for foot $i$ \\
    $\mathcal{F}$ 
    & Set of feet \\
    $\mathbf{F}_{i}$ 
    & Contact force of foot $i$\\
    $p_{z,b}$ 
    & Base height \\
    $h^*$ 
    & Target base height \\
    $\mathbf{g}=(g_x,g_y,g_z)$ 
    & Projected gravity vector in the base frame \\
    $T^a_i$ 
    & Current air time of foot $i$ \\
    $T^c_i$ 
    & Current contact time of foot $i$ \\

    \bottomrule
\end{longtable}

{\small
\renewcommand{\arraystretch}{1.25}

\begin{longtable}{p{0.30\textwidth} C{0.50\textwidth} C{0.15\textwidth}}
    \caption{\textbf{Reward terms} used in training. The equations show the unweighted reward terms, and $w$ denotes the configured reward weight.}
    \label{tab:rew_cfg}\\
    \toprule
    \textbf{Reward Term} & \textbf{Reward Equation} & \textbf{Weight} \\
    \midrule
    \endfirsthead

    \toprule
    \textbf{Reward Term} & \textbf{Reward Equation} & \textbf{Weight} \\
    \midrule
    \endhead

    \midrule
    \multicolumn{3}{r}{\textit{Continued on next page}} \\
    \endfoot

    \bottomrule
    \endlastfoot

    \multicolumn{3}{l}{\textbf{Manipulation rewards}} \\
    \midrule

    Grasp stability object retention $r_\mathrm{retention}$
    &
    $\displaystyle 0.5\left(-\tanh\!\left(\alpha(d-d_c)\right)-1.0\right),\quad $
    &
    $5.0$
    \\

    Grasp stability object slip $r_\mathrm{slip}$
    &
    $\displaystyle \mathbf{1}\!\left[\lVert \mathbf{v}_\mathrm{obj}-\mathbf{v}_\mathrm{ee}\rVert_2>v_c\right]$
    &
    $-0.3$
    \\

    Grasp stability force moderation $r_\mathrm{force}$
    &
    $\displaystyle \sum_{b\in \mathcal{B}_{g}}\lVert \mathbf{F}_{b}\rVert_2$
    &
    $-0.005$
    \\

    EE position exponential
    &
    $\displaystyle \exp\!\left(-\frac{2\lVert \mathbf{p}_{\mathrm{ee},\mathrm{cmd}}-\mathbf{p}_\mathrm{ee}\rVert_{1}}{\sigma^2}\right)$
    &
    $1.5$
    \\

    EE position L1
    &
    $\displaystyle -\lVert \mathbf{p}_{\mathrm{ee},\mathrm{cmd}}-\mathbf{p}_\mathrm{ee}\rVert_{1}$
    &
    $1.0$
    \\

    Manipulation action rate
    &
    $\displaystyle \sum_{j\in \mathbf{a}^{manip}}\left(a_{t,j}-a_{t-1,j}\right)^2$
    &
    $-0.002$
    \\

    Manipulation action smoothness
    &
    $\displaystyle \left\lVert \mathbf{a}^{manip}_{t}-\mathbf{a}^{manip}_{t-1}\right\rVert_2$
    &
    $-0.005$
    \\

    \midrule
    \multicolumn{3}{l}{\textbf{Locomotion rewards}} \\
    \midrule

    Base linear velocity
    &
    $\displaystyle \exp\!\left(-\frac{\lVert v_\mathrm{cmd}-v_{x,b}\rVert^2}{\sigma^2}\right)$
    &
    $1.5$
    \\

    Base yaw rate
    &
    $\displaystyle \exp\!\left(-\frac{\left(\omega_\mathrm{cmd}-\omega_{z,b}\right)^2}{\sigma^2}\right)$
    &
    $1.5$
    \\

    Base vertical velocity
    &
    $\displaystyle v_{z,b}^{2}$
    &
    $-2.5$
    \\

    Base roll/pitch rate
    &
    $\displaystyle \omega_{x,b}^{2}+\omega_{y,b}^{2}$
    &
    $-0.2$
    \\

    Leg joint torque
    &
    $\displaystyle \sum_{j\in \mathcal{J}_{all}}\tau_j^2$
    &
    $-2.0\times 10^{-5}$
    \\

    Leg joint acceleration
    &
    $\displaystyle \sum_{j\in \mathcal{J}_{all}}\ddot{q}_j^2$
    &
    $-2.5\times 10^{-7}$
    \\

    Feet air time
    &
    $\displaystyle \max\!\left(1-s\frac{\sum_i c_i(t_i-t^*)^2}{\max(\sum_i c_i,1)},0\right)$
    &
    $0.4$
    \\

    Feet contact
    &
    $\displaystyle \mathrm{clip}\!\left(\min_{i\in \mathcal{F}}\lVert \mathbf{F}_{i}\rVert_2,0,F_c\right)$
    &
    $0.005$
    \\

    Hip joint deviation
    &
    $\displaystyle \sum_{j\in \mathcal{J}_{hip}}\left|q_j-q^0_j\right|$
    &
    $-0.4$
    \\

    Thigh/calf joint deviation
    &
    $\displaystyle \sum_{j\in \mathcal{J}_{tc}}\left|q_j-q^0_j\right|$
    &
    $-0.04$
    \\

    Locomotion action rate
    &
    $\displaystyle \sum_{j\in \mathcal{A}^{loco}}\left(a_{t,j}-a_{t-1,j}\right)^2$
    &
    $-0.01$
    \\

    Locomotion action smoothness
    &
    $\displaystyle \left\lVert \mathbf{a}^{loco}_{t}-\mathbf{a}^{loco}_{t-1}\right\rVert_2$
    &
    $-0.02$
    \\

    Base height
    &
    $\displaystyle \left(p_{z,b}-h^*\right)^2$
    &
    $-5.0$
    \\

    Base orientation
    &
    $\displaystyle g_x^2+g_y^2$
    &
    $-1.0$
    \\

    Feet gait
    &
    $\displaystyle S(\mathcal{C}_1)S(\mathcal{C}_2)A(\mathcal{C}_\mathrm{fr})A(\mathcal{C}_\mathrm{re})A(\mathcal{C}_\mathrm{L})A(\mathcal{C}_\mathrm{R})$
    &
    $0.5$
    \\

\end{longtable}}

For the feet gait reward,
\begin{equation}
S(i,j)=\exp\!\left(
-\frac{
\mathrm{clip}((T^a_i-T^a_j)^2,0,c_m^2)
+
\mathrm{clip}((T^c_i-T^c_j)^2,0,c_m^2)
}{\sigma}
\right),
\end{equation}

\begin{equation}
A(i,j)=\exp\!\left(
-\frac{
\mathrm{clip}((T^a_i-T^c_j)^2,0,c_m^2)
+
\mathrm{clip}((T^c_i-T^a_j)^2,0,c_m^2)
}{\sigma}
\right),
\end{equation}
where $T^a_i$ and $T^c_i$ are the current air and contact times for foot $i$. The contact pairs are $\mathcal{C}_1=\{FR,RL\}$ and $\mathcal{C}_2=\{FL,RR\}$. Other foot pairs are defined as $\mathcal{C}_\mathrm{L}=\{FL,RL\}$, $\mathcal{C}_\mathrm{R}=\{FR,RR\}$, $\mathcal{C}_\mathrm{fr}=\{FR,FL\}$, $\mathcal{C}_\mathrm{re}=\{RL,RR\}$

\begin{table}[!htbp]
    \centering
    \caption{\textbf{Training hyperparameters.}}
    \label{tab:ppo_cfg}
    \begin{tabular}{lc}
        \toprule
        \textbf{Parameter} & \textbf{Value} \\
        \midrule
        Actor backbone dimension & $[256]$ \\
        Locomotion action head dimension & $[256,\ 128]$ \\
        Manipulation action head dimension & $[256,\ 128]$ \\
        Critic backbone dimension & $[256]$ \\
        Critic locomotion head dimension & $[256,\ 128,\ 64]$ \\
        Critic manipulation head dimension & $[256,\ 128,\ 64]$ \\
        Privileged info encoder dimension & $[256,\ 128]$ \\
        History encoder dimension & $[512,\ 256,\ 128]$ \\
        Privileged/history latent size & $96$ \\
        Tactile encoder dimension & $[128, 32]$ \\
        Tactile latent size & $32$ \\
        Initial learning rate & $0.001$ \\
        Learning rate schedule & Adaptive  \\
        Desired KL divergence & $0.01$ \\
        Clip range & $0.2$ \\
        Discount factor & $0.99$ \\
        GAE $\lambda$ & $0.95$ \\
        Number of environment steps per training batch & $24$ \\
        Learning epochs per training batch & $5$ \\
        Mini-batches per training batch & $4$ \\
        Initial advantage mixing ratio & $0$ \\
        Final advantage mixing ratio & $1$ \\
        Number of advantage mixing steps & $9000$ \\
        \bottomrule
    \end{tabular}
\end{table}

All networks are MLPs with ELU activation functions. The advantage mixing follows the form
\begin{align}
    J(\theta_\pi) &=
    \frac{1}{|\mathcal{D}|}
    \sum_{(s_t,a_t)\in\mathcal{D}}
    \left[
    \log \pi\left(a_t^{\mathrm{arm}} \mid s_t\right)
    \left(A^{\mathrm{manip}} + \beta A^{\mathrm{loco}}\right)
    +
    \log \pi\left(a_t^{\mathrm{leg}} \mid s_t\right)
    \left(\beta A^{\mathrm{manip}} + A^{\mathrm{loco}}\right)
    \right].
\end{align}
where $\beta$ is the advantage mixing ratio.~\citep{fu2022deep} provides a detailed definition and analysis. In our work, $\beta$ is increased linearly across the number of advantage mixing steps. 

\section{Deployment Setup}
We deploy the policy on a Unitree Go2 quadruped equipped with a back-mounted Interbotix WidowX 250 arm and tactile arrays on both inner gripper surfaces. Policy inference and observation processing run at $50~\mathrm{Hz}$ on an external PC, with Ethernet communication to the Go2 and USB communication to the arm and tactile sensors. Due to the limited torque capacity and bandwidth of the arm servos, the real-world evaluation uses a subset of the simulated end-effector trajectories to ensure reliable dynamic operation and hardware safety.

\end{document}